\documentclass[runningheads]{llncs}
\usepackage{graphicx}
\usepackage{amsmath} 
\usepackage{amsfonts} 

\begin{document}
\title{Brain-like approaches to unsupervised learning of hidden representations - a comparative study\thanks{Funding for the work is received from the Swedish e-Science Research Centre (SeRC), European Commission H2020 program, Grant Agreement No. 800999 (SAGE2), and Grant Agreement No. 801039 (EPiGRAM-HS). The simulations were performed on resources provided by Swedish National Infrastructure for Computing (SNIC) at the PDC Center for High Performance Computing, KTH Royal Institute of Technology.}}
%
%
\titlerunning{Brain-like approaches to unsupervised learning}

\author{Naresh Balaji Ravichandran\inst{1} \and
Anders Lansner\inst{1,2} \and
Pawel Herman\inst{1}}

\authorrunning{N. B. Ravichandran et al.}
%
\institute{Computational Brain Science Lab, KTH Royal Institute of Technology,\\ Stockholm, Sweden \and
Department of Mathematics, Stockholm University, Stockholm, Sweden\\
\email{\{nbrav,ala,paherman\}@kth.se}}
\maketitle              
\begin{abstract}
Unsupervised learning of hidden representations has been one of the most vibrant research directions in machine learning in recent years. In this work we study the brain-like Bayesian Confidence Propagating Neural Network (BCPNN) model, recently extended to extract sparse distributed high-dimensional representations. The usefulness and class-dependent separability of the hidden representations when trained on MNIST and Fashion-MNIST datasets is studied using an external linear classifier and compared with other unsupervised learning methods that include restricted Boltzmann machines and autoencoders.
\end{abstract}
\section{Introduction}
\label{sec:intro}

Artificial neural networks have made remarkable progress in supervised pattern recognition in recent years. In particular, deep neural networks have dominated the field largely due to their capability to discover hierarchies of hidden data representations. However, most deep learning methods rely extensively on supervised learning from labeled samples for extracting and tuning data representations. Given the abundance of unlabeled data there is an urgent demand for unsupervised or semi-supervised approaches to learning of hidden representations \cite{bengio2013representation}. Although early concepts of greedy layer-wise pretraining allow for exploiting unlabeled data, ultimately the application of deep pre-trained networks to pattern recognition problems rests on label-dependent end-to-end weight fine tuning \cite{erhan2009does}. At the same time, we observe a surge of interest in more brain plausible networks for unsupervised and semi-supervised learning problems that build on some fundamental principles of neural information processing in the brain \cite{illing2019biologically}. Most importantly, these brain-like computing approaches rely on local learning rules and label-independent biologically compatible mechanisms to build data representations whereas deep learning methods predominantly make use of error back-propagation (backprop) for learning the weights. Although backprop as a learning algorithm is highly efficient for finding good representations from data, there are several issues that make it an unlikely candidate model for synaptic plasticity in the brain. The most apparent issue is that the synaptic strength between two biological neurons is expected to comply with Hebb’s postulate, i.e. to depend only on the available local information provided by the activities of pre- and postsynaptic neurons. This is violated in backprop since synaptic weight updates need gradient signals to be communicated from distant output layers. Please refer to \cite{lillicrap2020backpropagation} for a detailed review of possible biologically plausible implementations of and alternatives to backprop. 

In this work we utilize the MNIST and Fashion-MNIST datasets to compare two classical representation learning networks, the autoencoder (AE) and the restricted Boltzmann machine (RBM), with two brain-like approaches to unsupervised learning of hidden representations, i.e. the recently proposed model by Krotov and Hopfield (KH) \cite{krotov2019unsupervised}, and the BCPNN model \cite{ravichandran2020learning}. In particular, we qualitatively examine the extracted hidden representations and quantify their class-dependent separability using a simple linear classifier on top of all the networks under investigation. This classification step is not part of the learning strategy, and we use it merely to evaluate the resulting representations.


Special emphasis is on the feedforward BCPNN model with a single hidden layer, which frames the update and learning steps of the neural network as probabilistic computations. BCPNN has previously been used in abstract models of associative memory \cite{lansner2009ann,sandberg2002bayesian}, action selection \cite{berthet2012action}, and in application to brain imaging \cite{benjaminsson2010novel} and data mining \cite{orre2000bayesian}. Spiking versions of BCPNN with biologically detailed Hebbian synaptic plasticity have also been developed to model different forms of cortical associative memory \cite{fiebig2017spiking,lundqvist2011theta,tully2014synaptic}. BCPNN architecture comprises many modules, referred to as hypercolumns (HCs), that in turn comprise a set of functional minicolumns (MCs) competing in a soft-winner-take-all manner. The abstract view of a HC in this cortical-like network is that it represents some attribute, e.g. edge orientation, in a discrete coded manner. A minicolumn conceptually represents one discrete value (a realization of the given attribute) and, as a biological parallel, it accounts for a local subnetwork of around a hundred recurrently connected neurons with similar receptive field properties \cite{mountcastle1997columnar}. Such an architecture was initially generalized from the primary visual cortex, but today has more support also from later experimental work and has been featured in spiking computational models of cortex \cite{lansner2009associative,rockland2010five}.

Finally, in this work we highlight an additional mechanism called structural plasticity, introduced recently to the BCPNN framework \cite{ravichandran2020learning}, which enables self-organization and unsupervised learning of hidden representations. Structural plasticity learns a set of sparse connections while simultaneously learning the weights of the connections. This is in line with structural plasticity found in the brain, where there is continuous formation and removal of synapses in an activity-dependent manner \cite{butz2009activity}. 

\section{Related Works}
\label{sec:related_works}

A popular unsupervised learning approach is to train a hidden layer to reproduce the input data as, for example, in AE and RBM. The AE and RBM networks trained with a single hidden layer are relevant here since learning weights of the input-to-hidden-layer connections relies on local gradients, and the representations can be stacked on top of each other to extract hierarchical features. However, stacked autoencoders and deep belief nets (stacked RBMs) have typically been used for pre-training procedures followed by end-to-end supervised fine-tuning (using backprop) \cite{erhan2009does}. The recently proposed KH model \cite{krotov2019unsupervised} addresses the problem of learning solely with local gradients by learning hidden representations only using an unsupervised method. In this network, the input-to-hidden connections are trained and additional (non-plastic) lateral inhibition provides competition within the hidden layer. For evaluating the representation, the weights are fixed, and a linear classifier trained with labels is used for the final classification. Our approach shares some common features with the KH model, e.g. learning hidden representations solely by unsupervised methods, and evaluating the representations by a separate classifier (\cite{illing2019biologically} provides an extensive review of methods with similar goals). 

All the aforementioned models employ either competition within the hidden layer (KH), or feedback connections from hidden to input (RBM and AE). The BCPNN uses only the feedforward connections, along with an implicit competition via a local softmax operation, the neural implementation of which would be lateral inhibition within a hypercolumn.

It is also observed that, for unsupervised learning, having sparse connectivity in the feedforward connections performs better than full connectivity \cite{illing2019biologically}. Even networks employing supervised learning, like convolutional neural networks (CNNs), force a fixed spatial filter to obtain this sparse connectivity. The BCPNN model takes an alternative adaptive approach by using structural plasticity to obtain a sparse connectivity.

\section{Bayesian Confidence Propagation Neural Network} 
\label{sec:BCPNN}

\begin{figure}[h]
\begin{center}
\includegraphics[width=\linewidth]{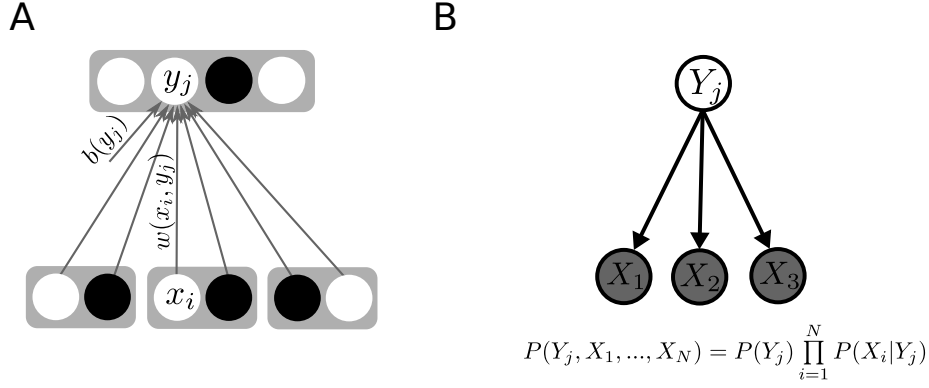}
\end{center}
\caption{Schematic of the BCPNN architecture with three input HCs and one hidden HC. {\bf A}. The neural network model where the input HCs are binary and the hidden HC is multinomial (gray boxes). The MCs within the HC are the discrete values of the hidden variable (open and shaded circles inside the box). {\bf B}. The equivalent probabilistic graphical model illustrating the generative process with each random variable (circle) representing a HC. The input HCs are observable (shaded circle) and the hidden HC is latent (open circle). The naive Bayes assumption renders the likelihood of generating inputs factorial.}
\label{figure-bcp}
\end{figure}

Here we describe the BCPNN network architecture and update rules \cite{lansner2009ann,ravichandran2020learning,sandberg2002bayesian}. The feedforward BCPNN architecture contains two layers, referred to as the input layer and hidden layer (Fig. \ref{figure-bcp}A). A layer consists of a set of HCs, each of which represents a discrete random variable $X_i$ (upper case). Each HC, in turn, is composed of a set of MCs representing a particular value $x_i$ (lower case) of $X_i$. The probability of $X_i$ is then a multinomial distribution, defined as $p(X_i=x_i)$, such that $\sum_{x_i} p(X_i=x_i) = 1$. In the neural network, the activity of the MC is interpreted as $p(X_i=x_i)$, and the activities of all the MCs inside a HC sum to one.

Since the network is a probabilistic graphical model (see Fig. \ref{figure-bcp}B), we can compute the posterior of a target HC in the hidden layer conditioned on all the source HCs in the input layer. We will use $x$’s and $y$’s to refer to the HCs in the input and hidden layers respectively. Computing the exact posterior $p(Y_j|X_1,...,X_N)$ over the target HC is intractable, since it scales exponentially with the number of units. The naive Bayes assumption $p(X_1,..,X_N|Y_j)= \prod_{i=1}^{N} p(X_i|Y_j)$ allows us to write the posterior as follows:
\begin{equation}
p(Y_j|X_1,...,X_N)
	= \frac{ p(Y_j)  \prod_{i=1}^{N} p(X_i|Y_j)}{p(X_1,...,X_N)}
	\propto p(Y_j) \prod_{i=1}^{N} p(X_i|Y_j)
\end{equation}

When the network is driven by input data $\{X_1,..,X_N\}=\{x_1^D,..,x_N^D\}$, we can write the posterior probabilities of a target MC in terms of the source MCs as:
\begin{equation}
p(y_j|x_1^D,...,x_N^D)
	\propto p(y_j) \prod_{i=1}^{N} p(x_i^D|y_j)
	= p(y_j) \prod_{i=1}^{N} \prod_{x_i} p(x_i|y_j)^{\mathbb{I}(x_i\!=\!x_i^D)} 	
\end{equation}

where $\mathbb{I}(\cdot)$ is the indicator function that equals 1 if its argument is true, and zero otherwise. We have written the posterior of the target MC as a function of all the source MCs (all $x_i$’s). The log posterior can be written as:   
\begin{equation}
\log p(y_j|x_1^D,...,x_N^D)
	\propto \log p(y_j) + \sum_{i=1}^{N} \sum_{x_i} \mathbb{I}(x_i\!=\!x_i^D) \log p(x_i|y_j)	
\end{equation}

Since the posterior is linear in the indicator function of data sample, $\mathbb{I}(x_i\!=\!x_i^D)$ can be approximated by its expected value defined as $\pi(x_i)=p(x_i\!=\!x_i^D)$. Except for $\pi(x_i)$, all the terms in the posterior are functions of the marginals $p(y_j)$ and $p(x_i,y_j)$. We define the terms bias $b(y_j)=\log p(y_j)$ and weight $w(x_i,y_j) = \log \frac{p(x_i,y_j)}{p(x_i)p(y_j)}$ in analogy with artificial neural networks. Note that in the weight term, we have added an additional $p(x_i)$ factor in the denominator to be consistent with previous BCPNN models \cite{sandberg2002bayesian,tully2014synaptic}, but this will not affect the computation since all terms independent of $y_j$ will absorbed in the activity normalization.

The inference step to calculate the posterior probabilities of the target MCs conditioned on the input sample is given by the activity update equations:
\begin{align}
h(y_j)
	&= b(y_j) + \sum_{i=1}^N \sum_{x_i} \pi(x_i) w(x_i,y_j) \\
\pi(y_j)
	&= \frac{\exp h(y_j) }{\sum_k \exp h(y_k) }
\end{align}

where $h(y_j)$ is the total input received by each target MC from which the posterior probability $\pi(y_j)=p(y_j|x_1^D,...,x_N^D)$ is recovered by softmax normalization of all MCs within the HC.

As each data sample is presented, the learning step incrementally updates the marginal probabilities, weights, and biases as follows:
\begin{align}
\tau_p \frac{d p(x_i)}{dt}
	&= \pi(x_i) - p(x_i) \\
\tau_p \frac{d p(x_i,y_j)}{dt}
	&= \pi(x_i) \: \pi(y_j) - p(x_i,y_j) \\
\tau_p \frac{d p(y_j)}{dt}
	&= \pi(y_j) - p(y_j) \\
b(y_j)
	&= \log \: p(y_j) \\
w(x_i,y_j)
	&= \log \frac{p(x_i,y_j)}{p(x_i)\:p(y_j)}
\end{align}
where the parameter $\tau_p$ is the learning time constant. The sets of Equations 4-5 and Equations 6-10 define the activity and learning update equations of the BCPNN architecture respectively. For the network with multiple input HCs and one hidden HC (as in Fig. \ref{figure-bcp}A), the computation is equivalent to a mixture model with each hidden MC representing one mixture component. For learning distributed representations from data, we can train multiple hidden HCs using the same principles. The network for unsupervised representation learning requires, in addition to the above computation, structural plasticity for learning a sparse set of connections from the input to hidden layer, which we discuss in detail in the next section. 

\subsection{Structual plasticity}

\begin{figure}
\begin{center}
\includegraphics[width=0.8\linewidth]{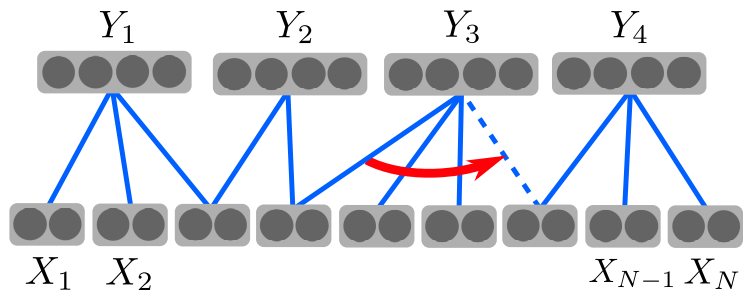}
\end{center}
\caption{ Structural plasticity. The input layer contains $N$ binary HCs, and the hidden layer contains four HCs (gray boxes). The existence of a connection between an input HC and hidden HC is shown as a blue line, i.e., $M_{ij}$=1. Structural plasticity involves each hidden HCs silencing an active connection and activating another silent connection. The red arrow shows one such flip operation for the third hidden HC.}
\label{figure-structplast}
\end{figure}

Structural plasticity builds a sparse set of connections from the input to hidden layer by iteratively improving from randomly initialized connections. We first define connections in terms of input HC to hidden HC, that is, when an input HC has an active connection to a hidden HC, all MCs within the input HC are connected to all MCs within the hidden HC. The connections are formulated as a connectivity matrix $M$, where $M_{ij}=1$ when the connection from the $i$th input HC to $j$th hidden HC is active, or $M_{ij}=0$ if silent\footnote{In analogy with biological synapses that can be non-existing, silent, or active, we adopt the term 'silent' for inactive connections.} (see Fig. \ref{figure-structplast}). Each $M_{ij}$ is initialized as active stochastically with probability $p_{ih}$, with $p_{ih}$ being the hyperparameter that controls the connection density. Once initialized, the total number of active incoming connections to each hidden HC is fixed whereas the outgoing connections from a source HC can be changed. For each connection, we compute the mutual information $I(X_i,Y_j)$ between the $i$th input HC and $j$th hidden HC and normalize by the number of active outgoing connections from the input HC to compute the score $\tilde{I}(X_i,Y_j)$:
\begin{equation}
\tilde{I}(X_i,Y_j)
	= \frac{I(X_i,Y_j)}{1+\sum_k M_{ik} } 
\end{equation}

Since the total number of active incoming connections is fixed, each hidden HC greedily maximizes the sum of score $\tilde{I}(X_i,Y_j)$ by silencing the active connection with the lowest score (change $M_{ij}$ from 1 to 0) and activating the silent connection with the highest score (change $M_{ij}$ from 0 to 1), provided that latter's score is higher than the former. We call this operation a flip and use a hyperparameter $n_{flips}$ to set the number of flips made per training epoch.

\section{Experiments}

Here we first describe the experimental setup for the BCPNN and three other related models for unsupervised learning as described in section 2. Next, we study qualitatively the hidden representations by examining the receptive fields formed by unsupervised learning. Finally, we provide quantitative evaluation of the representations learnt by the four models using class-dependent classification with a linear classifier.

\subsection{Data}

We ran the experiments on the MNIST \cite{lecun1998mnist} and Fashion-MNIST \cite{xiao2017fashion} datasets. Both datasets contain 60,000 training and 10,000 test samples of 28x28 grey-scale images. The images were flattened to 784 dimensions and the pixel intensities were normalized to the range [0,1]. The images acted as the input layer for the models and the labels were not used for unsupervised learning. 

\subsection{Models}

We considered four network architectures: BCPNN (c.f. section 3), AE, RBM and, KH. Each model had one hidden layer with 3000 hidden units. 

\subsubsection{BCPNN} The BCPNN network had an input layer with 784 HCs and 2 MCs per HC (pixel intensity was interpretted as probability of a binary variable) and a hidden layer with $n_{HC}$=30 and $n_{MC}$=100. The learning time constant was set as $\tau_p=60$ to roughly match the training time for one epoch (for details, see \cite{ravichandran2020learning}). The entire list of parameters and their values are listed in Table 1. The simulations were performed on code parallelized using MPI on a cluster of 2.3 GHz Xeon E5 processors and the training process took approximately fifteen minutes per run. 

\begin{table}
\begin{center}
\caption{BCPNN model parameters}\label{table:bcpnn-params}
\begin{tabular}{|l|l|l|}
\hline
Symbol &  Value & Descrption \\
\hline
$n_{epoch}$        & 5         & Number of epochs of unsupervised learning \\
$n_{HC}$           & 30        & Number of HCs in hidden layer \\
$n_{MC}$           & 100       & Number of MCs per HC in hidden layer \\
$\Delta t$         & 0.01   & Time-step \\
$\tau_p$  	      & 60       & Learning time-constant \\
$p_{ih}$     	      & 8\%       & Connectivity from input to hidden layer \\
$n_{flips}$        & 16         & Number of flips per epoch for structural plasticity \\
$\mu$              & 10        & Mean of poisson distribution for initializing MCs \\
\hline
\end{tabular}
\end{center}
\end{table}

\begin{figure}
\begin{center}
\includegraphics[width=\linewidth]{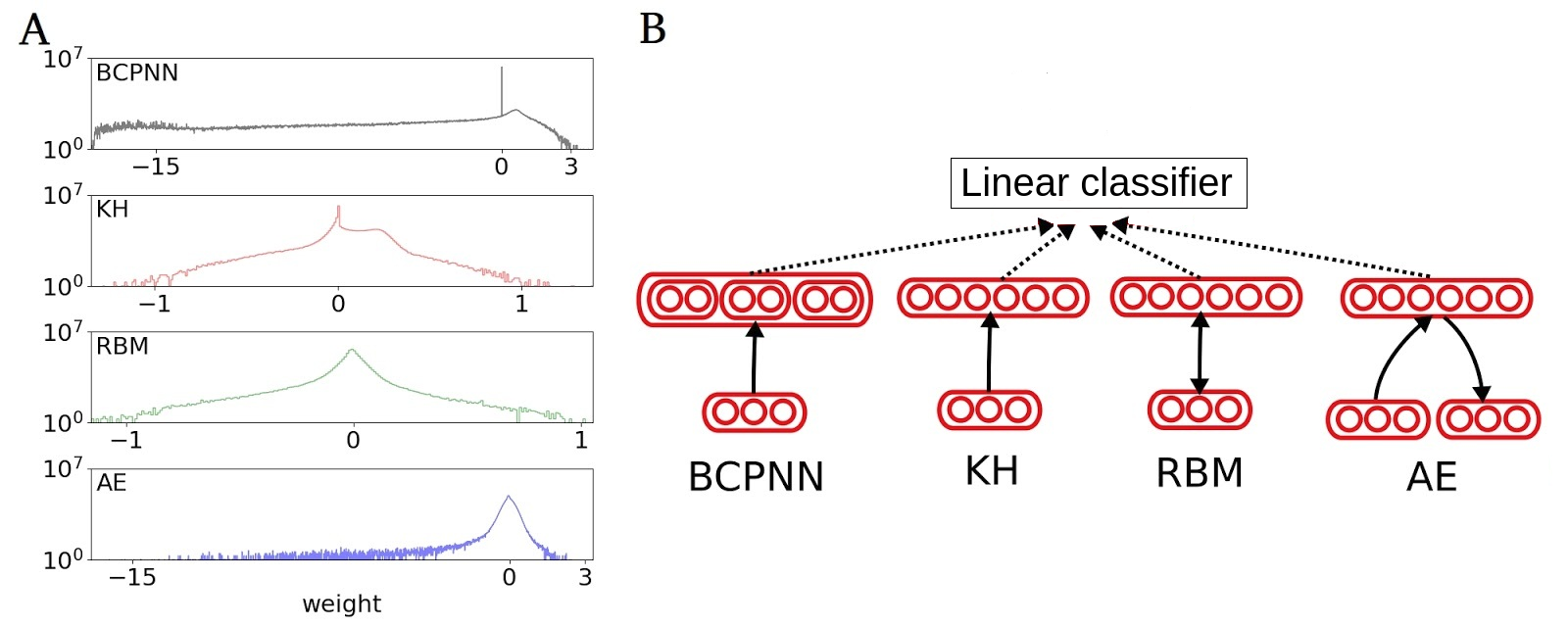}
\end{center}
\caption{ {\bf A}. Histogram of weights from the input layer to hidden layer. The horizontal axis has the minimum to maximum value of the weights as the range, and the vertical axis is in log scale. {\bf B}. Schematic of the four unsupervised learning models under comparison and the supervised classifier. The dotted lines imply we use the representations of the hidden layer as input for the classifier. }
\label{figure-classifiers}
\end{figure}

\subsubsection{KH} The KH network was reproduced from the original work using the code provided by Krotov and Hopfield \cite{krotov2019unsupervised}. We kept all the parameters as originally described, except for having 3000 hidden units instead of 2000, to be consistent in the comparison with other models.

\subsubsection{RBM} For the RBM network, we used sigmoidal units for both input and hidden layer. The weights were trained using the Contrastive Divergence algorithm with one iteration of Gibbs sampling (CD-1). The learning rate $\alpha$ was set as 0.01 and the training was done in minibatches of 256 samples for 300 epochs. 

\subsubsection{AE}  For the AE network, we used sigmoidal units for both hidden layer and reconstruction layer and two sets of weights, one for encoding from input to hidden layer and another for decoding from hidden to reconstruction layer. The weights were trained using the Adam optimizer and L2 reconstruction loss with an additional L1 sparsity loss on the hidden layer. The sparsity loss coefficient was determined by maximizing the accuracy of a held-out validation set of 10000 samples and set as $\lambda$=1e-7 for MNIST and $\lambda$=1e-5 for Fashion-MNIST. The training was in minibatches of 256 samples for 300 epochs.

\subsection{Receptive field comparison}

As can be observed in Fig. \ref{figure-classifiers}A, the distribution of weight values when trained on trained on MNIST considerably differs across the networks. It appears that the range of values for BCPNN corresponds to that reported for AE, whereas for KH and RBM, weights lie in a far narrower interval centered around 0. Importantly, BCPNN has by far the highest proportion of zero weights (90\%), which renders the connectivity truly sparse. 

\begin{figure}
\begin{center}
\includegraphics[width=\linewidth]{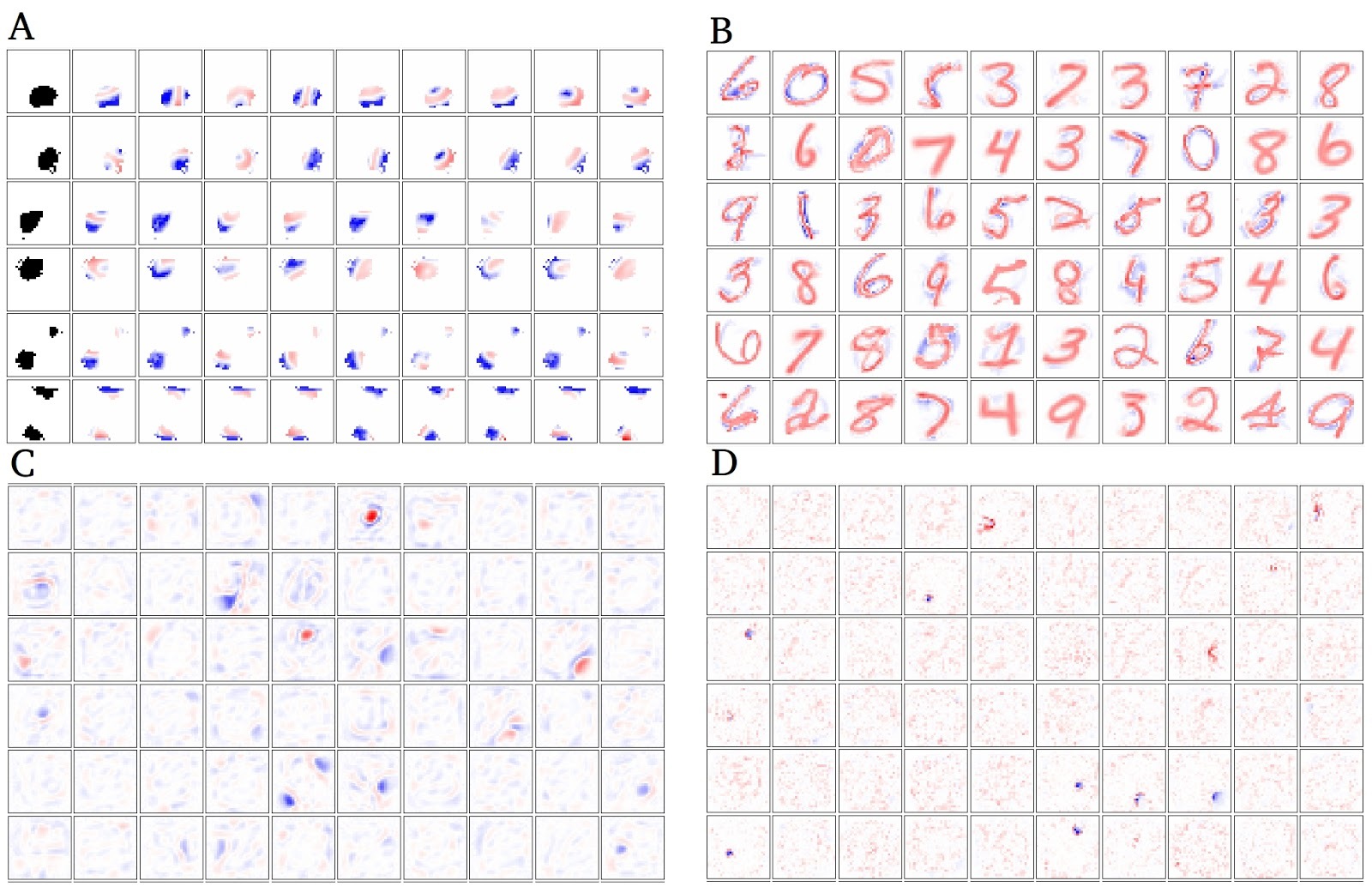}
\end{center}
\caption{ Receptive fields of different unsupervised learning methods. For each model, the positive and negative values are normalized, such that blue, white, and red represent the lowest, zero, and highest value of weights. {\bf A. BCPNN}: Each row corresponds to a randomly chosen HC and the constituent MCs of BCPNN. First column shows the receptive field of HC (black means $M_{ij}$=1). The remaining columns show the receptive field of nine randomly chosen MCs out of 100 MCs within the HC. {\bf B. KH}, {\bf C. RBM}, {\bf D. AE}: Receptive fields of 60 randomly chosen hidden units out of 3000.}
\label{figure-rfs}
\end{figure}

In Fig. \ref{figure-rfs}, we visualize the receptive fields of the four unsupervised learning networks trained on the MNIST dataset. Firstly, it is straightforward to see that the receptive fields of all the networks differ significantly. The RBM (Fig. \ref{figure-rfs}C) and AE (Fig. \ref{figure-rfs}D) have receptive fields that are highly localized and span the input space, a characteristic of distributed representations. The KH model (Fig. \ref{figure-rfs}B) has receptive fields that resemble the entire image, showing both positive and negative values over the image, as a result of Hebbian and anti-Hebbian learning \cite{krotov2019unsupervised}. Generally, local representations like mixture models and competitive learning, as opposed to distributed representations, tend to have receptive fields that resemble prototypical samples. With this distinction in mind, the receptive fields in the BCPNN should be closely examined (Fig. \ref{figure-rfs}A). The receptive fields of HCs (first column) are localized and span the input space, much like a distributed representation. Within each HC however, the MCs have receptive fields (each row) resembling prototypical samples, like diverse sets of lines and strokes. This suggests that the BCPNN representations are "hybrid", with the higher-level HCs coding a distributed representation, and the lower level MCs coding a local representation. 

\subsection{Classification performance}

For all the four models of unsupervised learning, we employed the same linear classifier for predicting the labels (see Fig. \ref{figure-classifiers}B). This allowed us to consistently evaluate the representations learned by the different models. The linear classifier considers the hidden layer representations as the input and the labels as the output. The output layer consists of softmax units for the 10 labels. The classifier’s weights were trained by stochastic gradient descent with the Adam optimizer using cross-entropy loss function. The training procedure used minibatches of 256 samples and a total of 300 training epochs.

\begin{table}
\caption{Accuracy comparison for MNIST and Fashion-MNIST datasets}
\label{table:bcpnn-params}
\begin{tabular}{|c|c|c|c|c|c|}
\hline
Model   & Hyperparameters & MNIST & MNIST & Fashion-MNIST & Fashion-MNIST \\
   &  & Train \% & Test \% & Train \% & Test \% \\
\hline 
BCPNN   & See Table 1 & 99.59 $\pm$ 0.01 & 98.31 $\pm$ 0.02  & 88.71 $\pm$ 0.05 & 86.31 $\pm$ 0.09 \\
KH
     &  See \cite{krotov2019unsupervised} & 98.75  $\pm$ 0.01  & 97.39 $\pm$ 0.06\footnotemark & 87.49 $\pm$ 0.03 & 85.10 $\pm$ 0.05 \\
RBM     & $\alpha=0.01$ & 98.92  $\pm$ 0.04  & 97.67 $\pm$ 0.10 & 88.13 $\pm$ 0.06 &  86.06 $\pm$ 0.12 \\
AE      & $\lambda=\{$1e-7,1e-5$\}$ & 100.0 $\pm$ 0.00  & 97.78 $\pm$ 0.09 & 88.52 $\pm$ 0.02 & 86.12 $\pm$ 0.08 \\
\hline
\end{tabular}
\end{table}

\footnotetext{This is lower than the test accuracy of 98.54\% reported by Krotov and Hopfield \cite{krotov2019unsupervised} who used a non-linear classifier with exponentiated ReLU activation and a non-linear loss function. Here we used instead a simpler linear classifier with softmax activation and cross-entropy loss. }

The results of the classification are shown in Table 2. All the results presented here are the mean and standard deviation of the classification accuracy over ten random runs. We performed three independent comparisons of BCPNN with KH, RBM, and AE using the Kruskal-Wallis test to evaluate statistical significance. On both MNIST and Fashion-MNIST, BCPNN outperforms KH, RBM, and AE (all with $p\!<\!0.0001$). 

For the BCPNN model, we set $n_{HC}$=30 and $n_{MC}$=100 in order to match the size of the hidden layer with the other models. However, BCPNN also scales well with the size of the hidden layer. From manual inspection, we found the best performance is at $n_{HC}$=200 and $n_{MC}$=100, where the test accuracy for MNIST is 98.58 $\pm$ 0.05 and Fashion-MNIST is 88.87 $\pm$ 0.08.

\section{Discussion}

We have evaluated four different network models that can perform unsupervised representation learning using biologically plausible local learning rules. We made our assessment relying on the assumption that the usefulness of representations is reflected in their class separability, which can be quantified by classification performance, similar to recent unsupervised learning methods \cite{oord2018representation}. Learning representations without supervised fine-tuning is a harder task compared to similar networks with end-to-end backprop, since the information about the samples’ corresponding labels cannot be utilized. Additionally, our unsupervised learning method relies on local Hebbian-like learning without any global optimisation criterion as in backprop learning, which makes the task even harder. Yet, we show that the investigated BCPNN model performs well, comparable to the 98.5\% accuracy of multi-layer perceptron networks on MNIST with one hidden layer trained using end-to-end backprop \cite{lecun1998gradient}.

We also showed that the recently proposed BCPNN model performs competitively against other unsupervised learning models. The modular structure of the BCPNN layer led to “hybrid” representations that differ from the well-known distributed and local representations. In contrast to the minibatch method of other unsupervised learning methods, learning in BCPNN was chosen to remain incremental using dynamical equations, since such learning is biologically feasible and useful in many autonomous engineering solutions. Despite the slow convergence properties of an incremental approach, BCPNN required only 5 epochs of unsupervised training, in comparison to 300 epochs for AE and RBM, and 1000 epochs for KH. The incremental learning, along with modular architecture, sparse connectivity, and scalability of BCPNN is currently also taken advantage of in dedicated VLSI design \cite{stathis2020ebrainii}.

One important difference between current deep learning architectures and the brain concerns the abundance of recurrent connections in the latter. Deep learning architectures rely predominantly on feedforward connectivity. A typical cortical area receives only around 10\% of synapses from lower order structures, e.g. thalamus, and the rest from other cortical areas \cite{douglas2007recurrent}. These feedback and recurrent cortical connections are likely involved in associative memory, constraint-satisfaction e.g. for figure-ground segmentation, top-down modulation and selective attention \cite{douglas2007recurrent}. Incorporating these important aspects of cortical computation can play a key role in improving our machine learning models and approaches.

It is important to note that the unsupervised learning models discussed in this work are proof-of-concept designs and not meant to directly model some specific biological system or structure. Yet, they may shed some light on the hierarchical functional organization of e.g. sensory processing streams in the brain. Further work will focus on extending our study to multi-layer architectures.

%
%
%
%

\bibliographystyle{splncs04}
\bibliography{main}

\end{document}